\documentclass[sigconf]{acmart}

\AtBeginDocument{%
  \providecommand\BibTeX{{%
    \normalfont B\kern-0.5em{\scshape i\kern-0.25em b}\kern-0.8em\TeX}}}

\begin{document}

\title{Simulations of MRI Guided and Powered Ferric Applicators for Tetherless Delivery of Therapeutic Interventions}

\author{Wenhui Chu}
\email{wchu@uh.edu}
\affiliation{%
  \institution{MRI Lab, University of Houston}
  \city{Houston}
  \state{TX}
  \country{USA}
}
\author{Khang Tran}
\email{kqtran8@uh.edu}
\affiliation{%
  \institution{MRI Lab, University of Houston}
  \city{Houston}
  \state{TX}
  \country{USA}
}

\author{Nikolaos V. Tsekos}
\email{nvtsekos@central.uh.edu}
\affiliation{%
  \institution{MRI Lab, University of Houston}
  \city{Houston}
  \state{TX}
  \country{USA}
}

\renewcommand{\shortauthors}{Wenhui Chu and Nikolaos V. Tsekos}

\begin{abstract}

Magnetic Resonance Imaging (MRI) is a well-established modality for pre-operative planning and is also explored for intra-operative guidance of procedures such as intravascular interventions. Among the experimental robot-assisted technologies, the magnetic field gradients of the MRI scanner are used to power and maneuver ferromagnetic applicators for accessing sites in the patient's body via the vascular network. In this work, we propose a computational platform for preoperative planning and modeling of MRI-powered applicators inside blood vessels. This platform was implemented as a two-way data and command pipeline that links the MRI scanner, the computational core, and the operator. The platform first processes multi-slice MR data to extract the vascular bed and then fits a virtual corridor inside the vessel. This corridor serves as a virtual fixture (VF), a forbidden region for the applicators to avoid vessel perforation or collision. The geometric features of the vessel centerline, the VF, and MRI safety compliance (dB/dt, max available gradient) are then used to generate magnetic field gradient waveforms. Different blood flow profiles can be user-selected, and those parameters are used for modeling the applicator's maneuvering. The modeling module further generates cues about whether the selected vascular path can be safely maneuvered. Given future experimental studies that require a real-time operation, the platform was implemented on the Qt framework (C/C++) with software modules performing specific tasks running on dedicated threads: PID controller, generation of VF, generation of MR gradient waveforms.  
\end{abstract}



\maketitle

\section{Introduction}

Minimally invasive surgery (MIS) encompasses a specialized type of surgical method that allows surgeons to dramatically reduce patient discomfort and recovery time. Many researchers take efforts in micro-and nanorobotics, which have evolved to encompass platforms and technologies aimed at improving interventional procedures in the human body through the use of navigable or actuated magnetic spheres. The advent of real-time image guidance provides new opportunities in the field of interventional medicine [1]. We want to combine such magnetic actuation with a proper imaging modality, allowing for the tracking of MRbot in the body and opening the possibility for closed-loop servo control along pre-planned trajectories. The smaller the MRbot is, the wider the operating range becomes through access to the blood vessels such as capillaries. The use of MRbot in MIS medicine is becoming a trend in robotics research and different approaches have been investigated.

Magnetism is a prioritized choice for moving such microscale robots [2]-[4]. The ferromagnetic core embedded in the micro-device will provide the driving force from an MRI system. It is difficult to offer minimally invasive surgery, because of continuous cardiac motion. So in our work, we used a method which is called magnetic resonance targeting (MRT), which has been demonstrated in vivo [24]. It is based on magnetic actuation and is known as Magnetic Resonance Navigation (MRN). MRN has drug-loaded microcarriers embedded in a magnetic sphere, which moves through the vessel network from the injection site to the targeted region. The targeting of therapeutics to tumors has a great advantage in increasing the efficiency of treatments while reducing their secondary toxicity effects [5].

\section{METHODOLOGY}

We use the vessel centerline data extracted from the 3D vascular model produced by 3D angiography images. In the present work, we use MRI to generate a safe zone, which is called a safe corridor, in real-time inside the heart. The goal of the present paper is to simulate the movement of a sphere through 3D models of vessel corridors. Because the human blood circulatory system is made of arteries, veins, and capillaries, the respective diameters of the pathways through which a biomedical robot must pass range from tens of micrometers to several centimeters. It becomes obvious that being able to reduce the size of such an MRbot would allow a larger percentage of locations in the human body to be reached. In many cases, an untethered implementation is suitable in order to propel MRbots into the human blood circulatory system, and it is obvious that propelling such wireless MRbots in the human cardiovascular system with existing technologies is a great technical challenge [6]. 

In order to propel such MRbots into the body fluids, especially in the human blood circulatory system, it must be considered part of the normal blood flow. The smaller these MRbots are, the wider the operating range becomes through access to the blood vessels such as capillaries.

Our goal is to control and propel the micro-robot by magnetic field gradients inside the blood vessels. We use a proportional integral derivative (PID) controller to correct and stabilize the trajectory when the sphere moves in more tortuous segments of the vessel. The physical parameters of the blood flow and the diameters of the blood vessels that are targeted determine the drag force and the constraints experienced by the MRbots. The quantity of the ferromagnetic particles is performed and determined by the medical application.

\section{THEORETICAL MODEL}

According to the present invention, a homogeneous spherical MRbot is considered in this paper. Since the gradients are generated uniformly throughout the border of the scanner, forces can be produced everywhere inside the body and can be precisely controlled with respect to time thus making it the ideal solution. In small blood vessels, the buoyancy of the MRbots and their weight can be neglected because such parameters are smaller than the drag force by orders of magnitude. The blood flow in the arterial system is pulsatile and much faster at the exit of the heart. The blood velocity decreases away from the heart to a cm/s then mm/s [7].

\subsection{Interpolation}

Interpolation is an important reference for medical imaging applications. In the volumetric imaging of this paper, we use an interpolation method called Piecewise Cubic Hermite Polynomials (pchip) to compensate for nonhomogeneous data sampling. Given an interval [a, b], a function d : [a,b] $\to$ $\Re$ with derivative d' : [a,b] $\to$ $\Re$, we use a cubic hermite spline s that approximates d over [a,b]. Our method will use the nearest method to split the interval [a,b] into N subintervals of nodes $\overrightarrow{x}$ = ($x_0$,$x_1$,$x_2$,...,$x_N$) with a = $x_0$$<$$x_1$$<$$x_2$$<$...$<$$x_N$ = b. In order to calculate 3D interpolation, we introduce a new parameter T as a tuple (pathDistance, x), where pathDistance equals the distance from the start point to the current point. 

\subsection{ Tortuosity}
Tortuosity is defined as the minimal geometrical path in a media which may require the computation of the minimal geodesic trajectories. At this point, we indicate that tortuosity is a directional parameter. It is a parametrically-defined space curve in three dimensions given in Cartesian coordinates by $\mathnormal{K}$ = ($\mathnormal{x}$(t), $\mathnormal{y}$(t), $\mathnormal{z}$(t)), corresponding to the three Cartesian axes ($\mathnormal{X}$, $\mathnormal{Y}$, and $\mathnormal{Z}$).   
\begin{equation}
 \label{e2}
K =\frac{\sqrt{(z''y'-y''z')^2+(x''z'-z''x')^2+(y''x'-x''y')^2}}{{(x'^2+y'^2+z'^2)}^{\frac{3}{2}}}\qquad
\end{equation}

where $x'$, $y'$, $z'$ are the first derivative of the desired path, $x''$, $y''$, $z''$ are the second derivative of the desired path.

\subsection{Blood Velocity} 
Pulsation arises from the pumping action of the heart, which forces a pulsating blood flow into the arteries, thereby creating a time-varying pressure that acts on the vessel wall [8]. The models are zero-dimensional (0D) models of the circulation, to one-dimensional (1D) models of blood pressure and flow propagation [9], to three-dimensional (3D) fluid-structure interaction techniques. Each approach has its own merits and limitations. We use 3D methods to represent complex flows, wave propagation, and blood flow–vessel wall interactions. The drag force produced by the blood acting on a sphere in an infinite extent of fluid can be calculated as:
\begin{equation}
 \label{e1}
F_{drag}=\frac{1}{2} \cdot C_d\cdot p\cdot Re \cdot 
\left\vert V_{blood} - V_s\right\vert  
\end{equation}
where $C_d$ is the drag coefficient, $\mathnormal{p}$ is the blood density (1.025 $kg/{m}^3$), $R_e$ is the reference area.

\subsubsection{Different Type of Blood Flow} 
In the present work, we studied the blood flow in the three types of pulsative: constant, normal, and high heart rate flow [10]. The gradients are updated every 100ms but the position of the sphere can be found every Tp millisecond. We use different Tp values in the different flows to test and show the results in Qt platform. 

\begin{figure*}[h]
\centering
   \includegraphics[height=3in,width=7in]{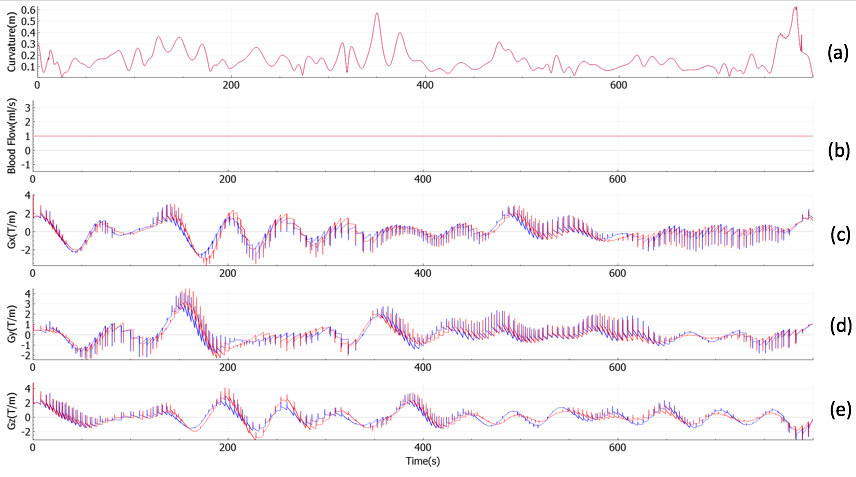}
    \caption{Results of Tp = 100ms (blue) and Tp = 200ms (red) of the simulation in the constant blood flow. (a) is a plot of the curvature of the path. (b) shows the constant blood flow. (c), (d) and (e) present the gradients generated by the MRI scanner for each axis.}\label{GC}
\end{figure*}

\begin{figure*}[h]
\centering
   \includegraphics[height=3in,width=7in]{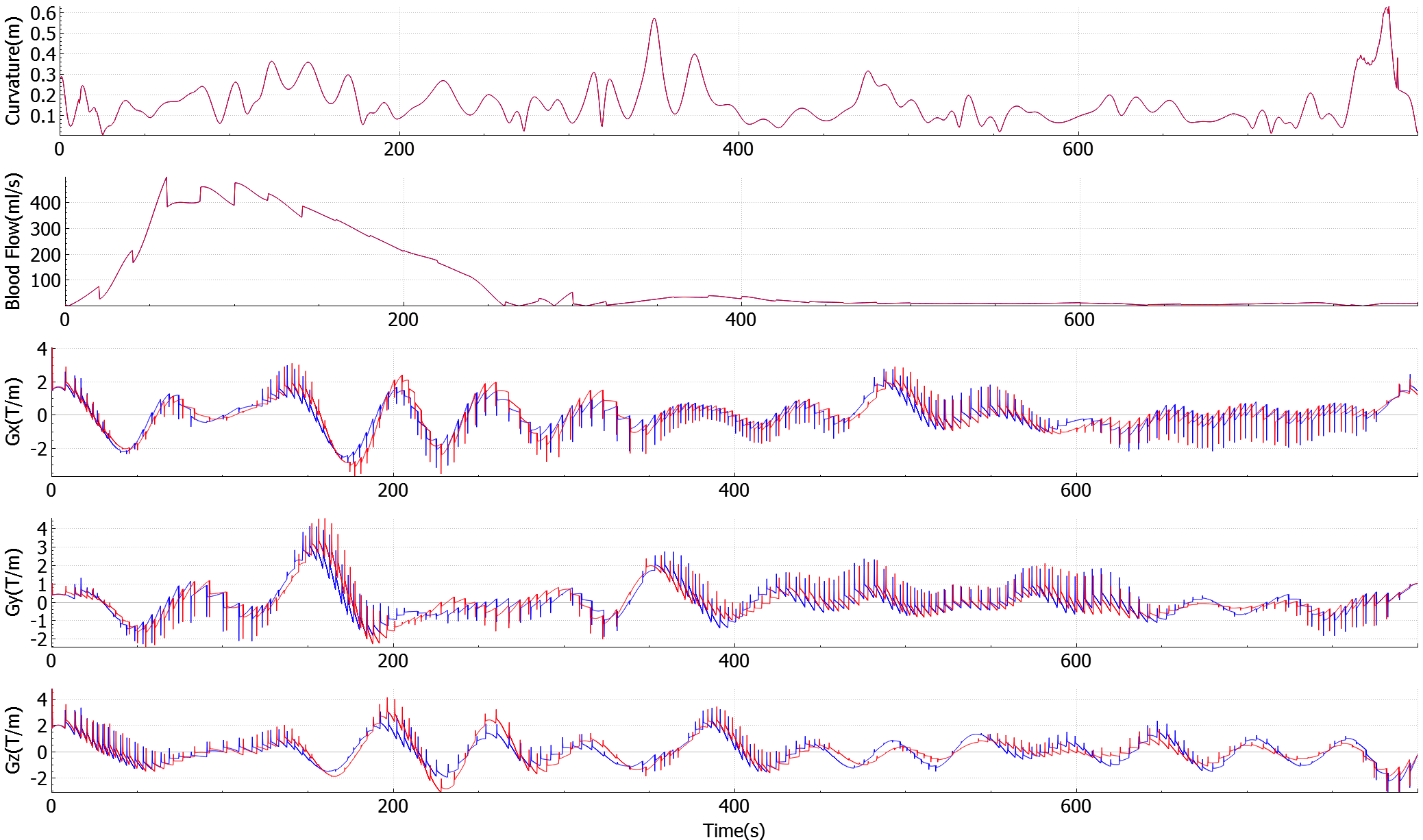}
    \caption{Results of Tp = 100ms (blue) and Tp = 200ms (red) of the simulation in the normal blood flow. (a) is a plot of the curvature of the path. (b) shows the normal blood flow. (c), (d) and (e) present the gradients generated by the MRI scanner for each axis.}\label{Ga}
\end{figure*}

\begin{figure*}
\centering
   \includegraphics[height=3in,width=7in]{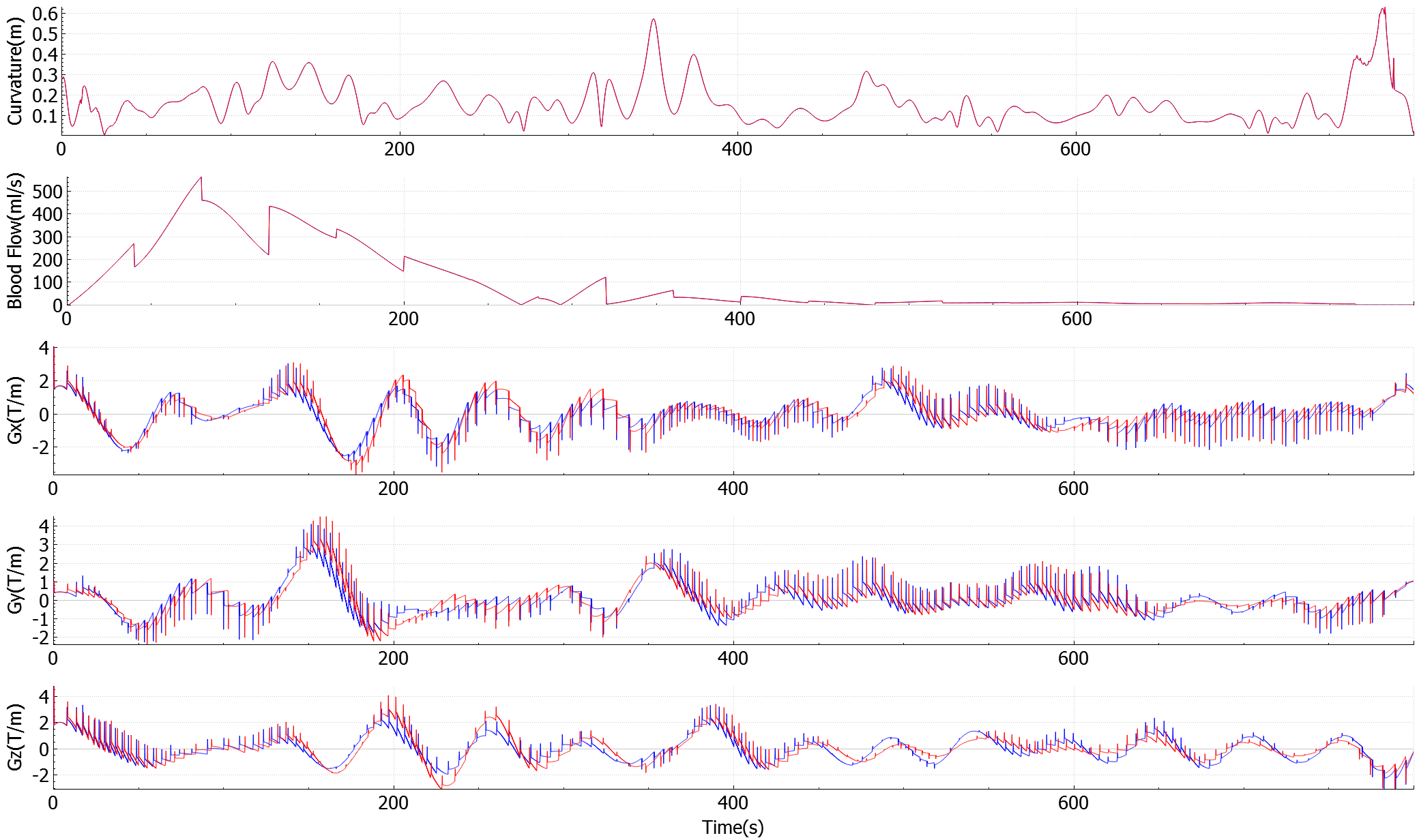}
    \caption{Results of Tp = 100ms (blue) and Tp = 200ms (red) of the simulation in fast heart rate of blood flow. (a) is a plot of the curvature of the path. (b) shows the fast heart rate blood flow. (c), (d) and (e) present the gradients generated by the MRI scanner for each axis.}\label{Gh}
\end{figure*}

\subsection{Velocity of Sphere} 
The trajectory generates the desired velocity V(r) along with the track P(r), counting its vector for each one of the points of the path [11][12].  

\begin{equation}
 \label{e2}
V(r) = \frac{V_0}{1+K/K_0}+\frac{R_s-R_{GC}}{{R_{0}}}\\
\end{equation}

where $R_s$ is the radius of the MRbot, $V_0$, $K_0$, and $R_0$ are constants allowing for adjustment of the velocity profile.

\subsection{Overview of MRI gradients} 
In order to maneuver safely, we introduce virtual fixtures (VF) which are virtual software-based constraints. There are two main parts with which we are concerned: one is continuous tracking and path imaging, the other one is maneuvering safely. Several means of propulsion are proposed to be embedded onto such an MRbot [13][14]. We propose a trajectory plan, which aims to follow a planning path to ensure safety and to control the MRbots. We use VF to generate virtual guidance cues for directing MRbot inside the vessels. When we use the guidance cues, the MRbot should reach the destination point accurately and not hurt the vessels. In the proposed approach, we simulate MRI Guided and Powered Ferric Applicators for Tetherless Delivery of Therapeutic Interventions. This is particularly true for a given magnitude of magnetic gradients generated by the MRI system. The size of the ferromagnetic body must be selected according to the diameters of the blood vessels, which is then used for the predefined path to the final target location.

\subsubsection{Control Model}
It is important to know the trajectory, the position of the MRbot, and gradients. Three set points need to be generated. A setpoint that contains the coordinate of points on the path centerline $P_t$ needs to be generated [15]. The velocity of the setpoint for the sphere is taken as the reference value. The trajectory controller is composed of a PID regulator and a feedforward component that directly outputs the optimal control. Errors on the velocity can be calculated with :

\begin{equation}
 \label{e1}
Error_v = V_c-V_s + K_r * (P_c-P_s)
\end{equation}

where $V_c$ is current velocity, $V_s$ is velocity setpoint, and Kr = 0.7 which defines how much error in position is corrected compared to the error in the velocity. The last two variables are $P_c$, which is current position, and $P_s$, which is the position setpoint.   

The PID regulator :
\begin{equation}
 \label{e1}
PF= -k_p* Error_v
\end{equation}

\begin{equation}
 \label{e1}
PI= PI-(Error_v* \delta *k_i)
\end{equation}

\begin{equation}
 \label{e1}
PD= -k_d* Error_{dt}
\end{equation} 

\begin{equation}
 \label{e1}
Error_{dt} = (Error_v - Error_p) / \delta
\end{equation} 
The parameters used for the PID controller are $k_p$ = 2, $k_i$ = 1, $k_d$ = 0.01, $\delta $ = base velocity, and $Error_p$ is errors on the previous position. 

\begin{equation}
 \label{e1}
Moment_s = M*Vol
\end{equation} 

\begin{equation}
 \label{e1}
Vol = \frac{4}{3}\times \pi \times r^3
\end{equation} 
where $Moment_s$ is the moment sphere, $\mathnormal{M}$ $(1.9496 \times 10^6$ A/m) is the magnetization sphere, $\mathnormal{Vol}$ is the volume sphere, and $\mathnormal{r}$ (0.3 mm) is the sphere radius. 

$\mathnormal{FF}$ is the optimal control that corresponds to the gradients of compensation for the drag force. It is produced by the blood:

\begin{equation}
 \label{e1}
FF=  \frac{1}{2} \cdot C_d \cdot p\cdot R_e \cdot V
\end{equation} 

where $C_{d}$ = 0.47 is the drag coefficient, $\mathnormal{p}$ is blood density, $\mathnormal{R_e}$ is the frontal area of the sphere, and $\mathnormal{V}$ is the blood velocity.

\subsubsection{Gradients Control}
Gradients ($\mathnormal{G}$) can be calculated with:
\begin{equation}
 \label{e1}
G= (1/Moment_s)*(PF+PI+PD+FF)
\end{equation} 

where $Moment_s$ is the moment sphere, $\mathnormal{PF}$, $\mathnormal{PI}$, $\mathnormal{PD}$ are PID controllers, and $\mathnormal{FF}$ is the optimal control. See Figure ~\ref{GC},~\ref{Ga},~\ref{Gh}. PF will react immediately to an error value, and try to adjust the sphere's location close to the setpoint. The higher PF, the faster reaction of the controller. PI will accumulate over time until the sphere move to the set point which means the longer it takes to reach the setpoint, the more PI will influence the output. PD means how fast the process value is reaching the setpoint.  From the figures, we can see that there are some spikes in the gradients. There is one reason that may result in spikes. Because PID is a feedback system, PI always needs an initialization value to calculate the first PI, even if we avoid the case where i = 0, which leads to the appearance of spikes. 

\subsubsection{Magnetic Force}
In order to get a strong magnetic force, the parameters to be maximized are the magnetization of the MRbot. Knowing that a ferromagnetic propulsion core is easy to fabricate and miniaturize as the viscosity of the blood is lower than heart tissues, the force is directly proportional to the ferromagnetic volume. Although the diameter of the blood vessels may vary due to cycles of heart contraction and the blood vessels, which affect drag force, our model considered the blood vessels as rigid cylindrical tubes.
\begin{equation}
 \label{e1}
 F=M \cdot G \cdot {V_{o}}
\end{equation}

Magnetic Resonance Propulsion (MRP) consists of applying magnetic gradients to exert a displacement force on MRbot. In (13), F is the magnetic force (N) produced by magnetic gradients (T $m^ {-1}$), M is the magnetization of the material (A/m), ${V_{o}}$ is the volume of the ferromagnetic body ($m^3$), G is the gradient or spatial variation of the magnetic induction. It can be seen from the equation that as the volume or the overall dimension of a ferromagnetic core in an MRbots or the volume of the ferromagnetic MRbots itself decreases, the force induced by the MRI system will decrease as well [16][17]. In order to produce sufficient magnetic force, using a ferromagnetic material with a stronger saturation magnetization is required. Friction between the sphere and tube walls is not considered in this study as gradients which are provided by MRI systems are strong enough to levitate MRbots.

\begin{figure*}[ht!]
\centering
   \includegraphics[height=3in,width=7in]{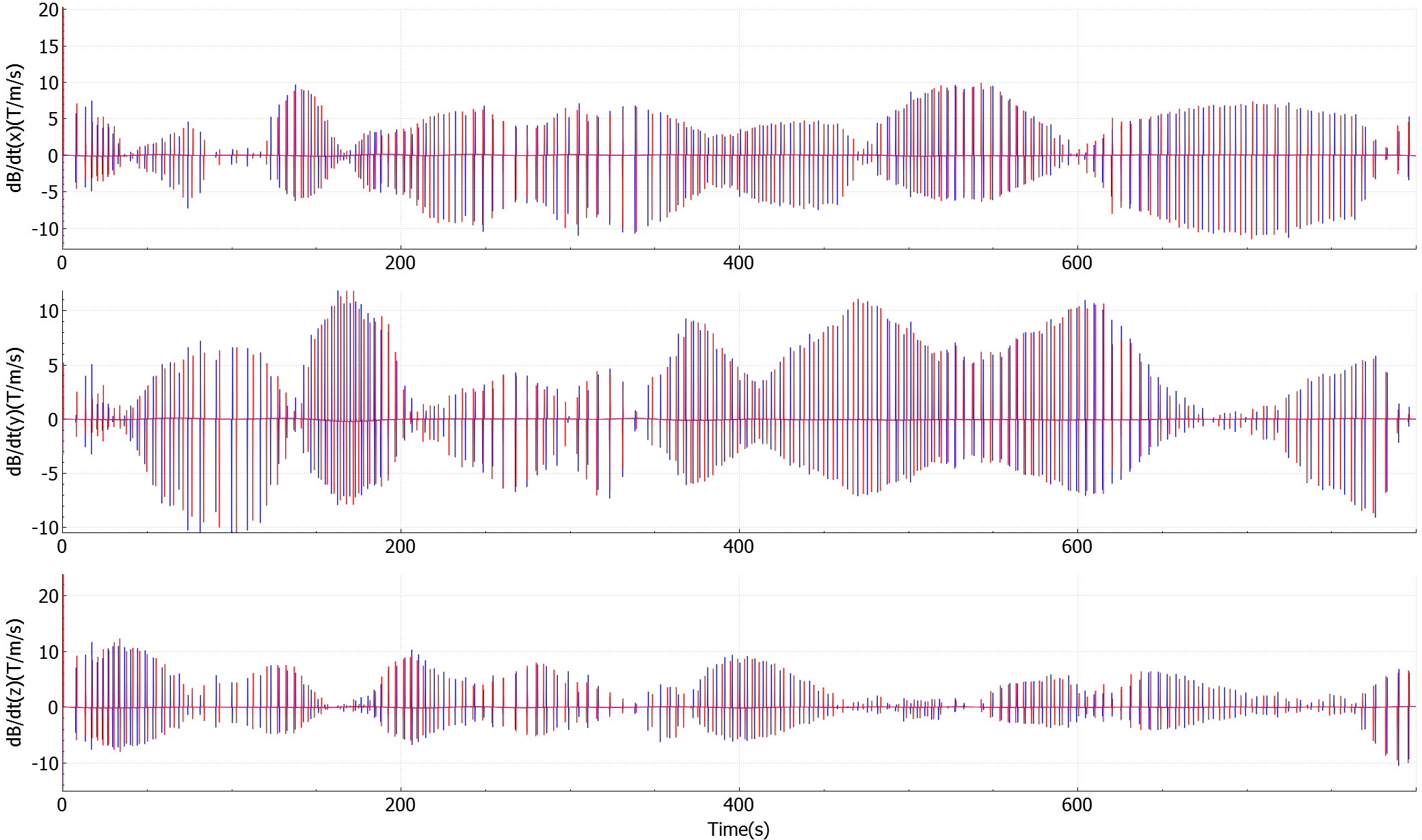}
    \caption{Results of Tp = 100ms (blue) and Tp = 200ms (red) of the simulation in the constant blood flow. (a), (b) and (c) show dB/dt for each axis. The blood flow is 1ml/s.}\label{dC}
\end{figure*}

\begin{figure*}[h]
\centering
   \includegraphics[height=3in,width=7in]{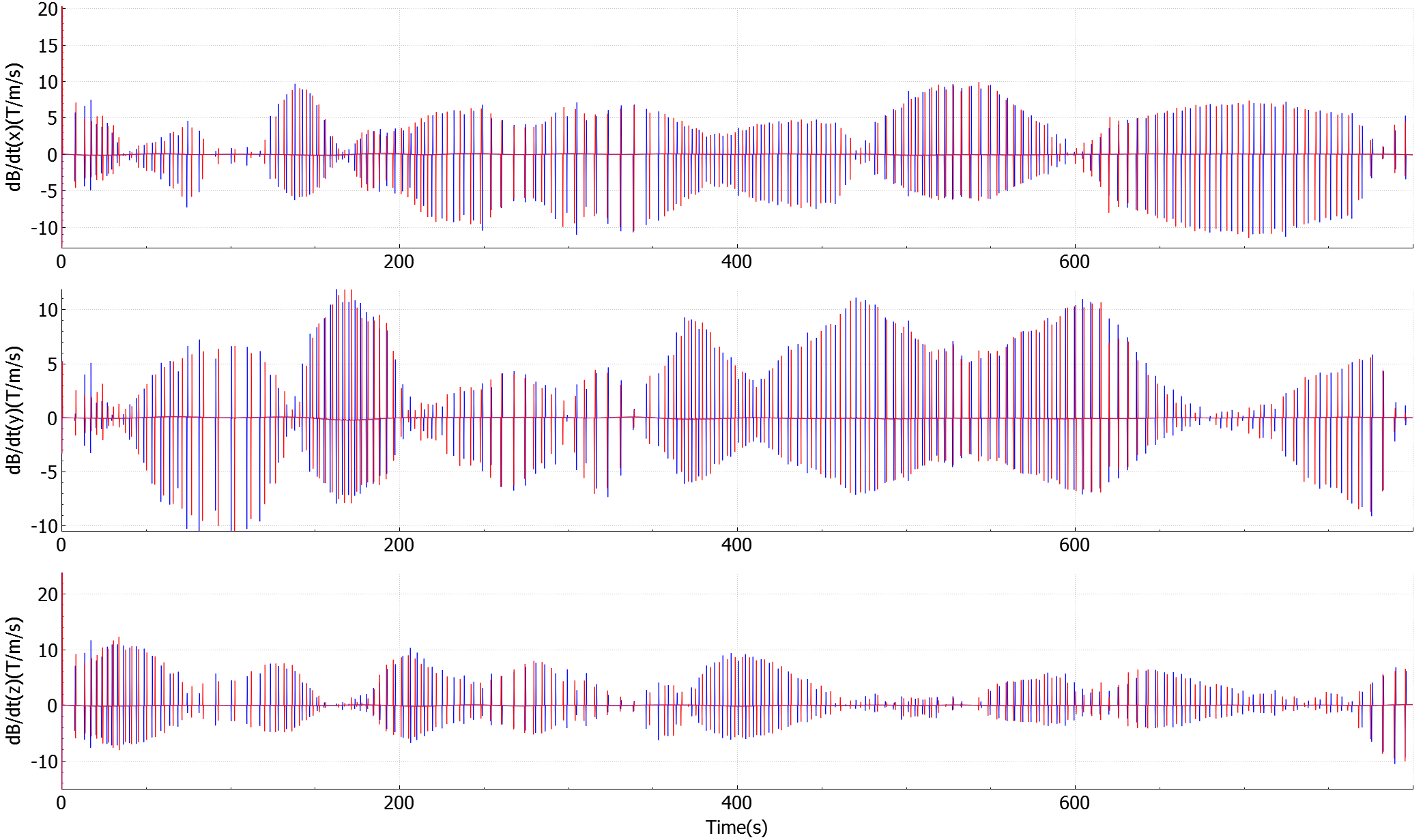}
    \caption{Results of Tp = 100ms (blue) and Tp = 200ms (red) of the simulation in the normal blood flow. (a), (b) and (c) show dB/dt for each axis.}\label{Da}
\end{figure*}

\begin{figure*}[h]
\centering
   \includegraphics[height=3in,width=7in]{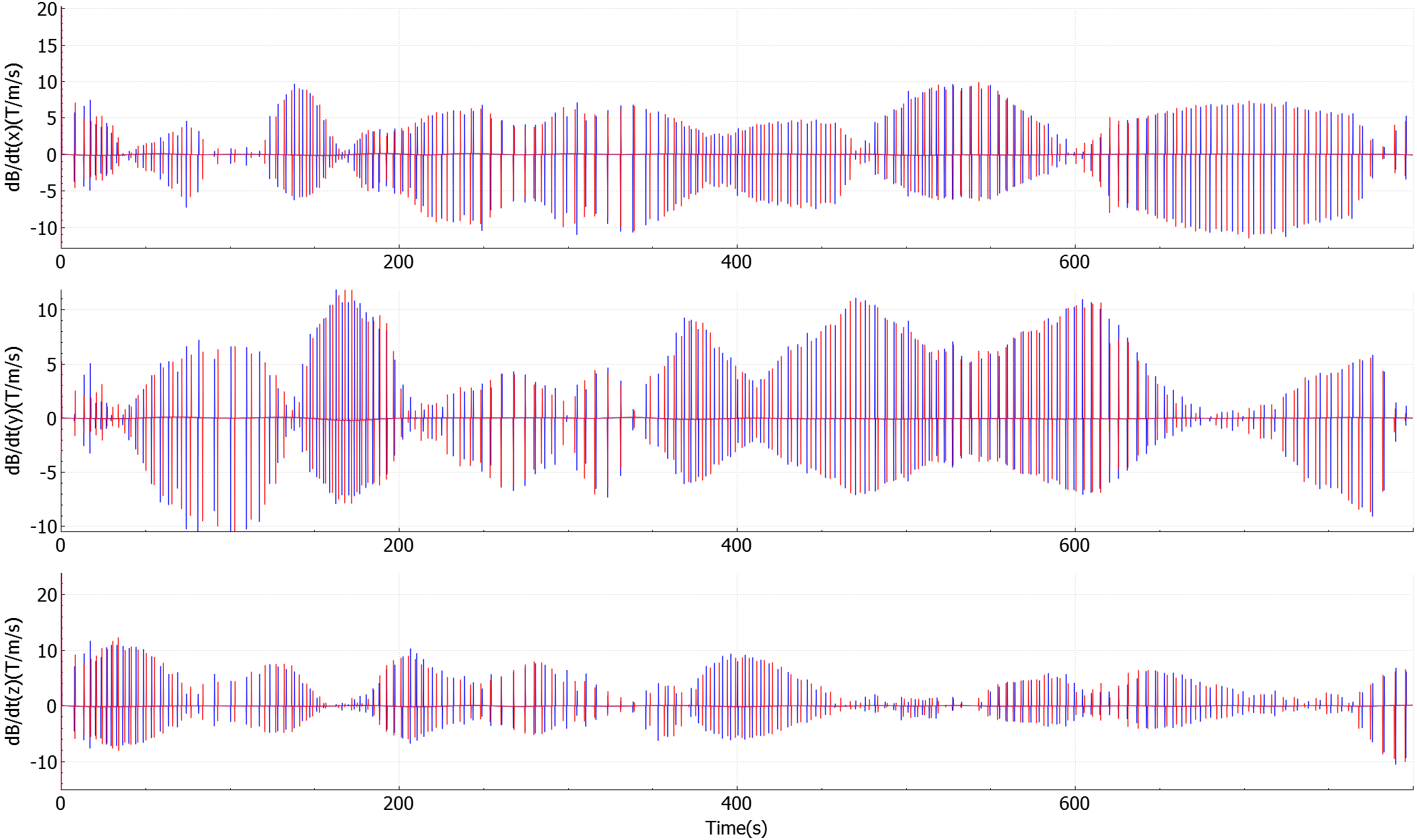}
    \caption{Results of Tp = 100ms (blue) and Tp = 200ms (red) of the simulation in the fast heart rate of blood flow. (a), (b) and (c) show dB/dt for each axis.}\label{Dh}
\end{figure*}

\section{SIMULATION AND RESULTS}

\begin{figure*}[h]
\centering
\includegraphics[width = 9cm]{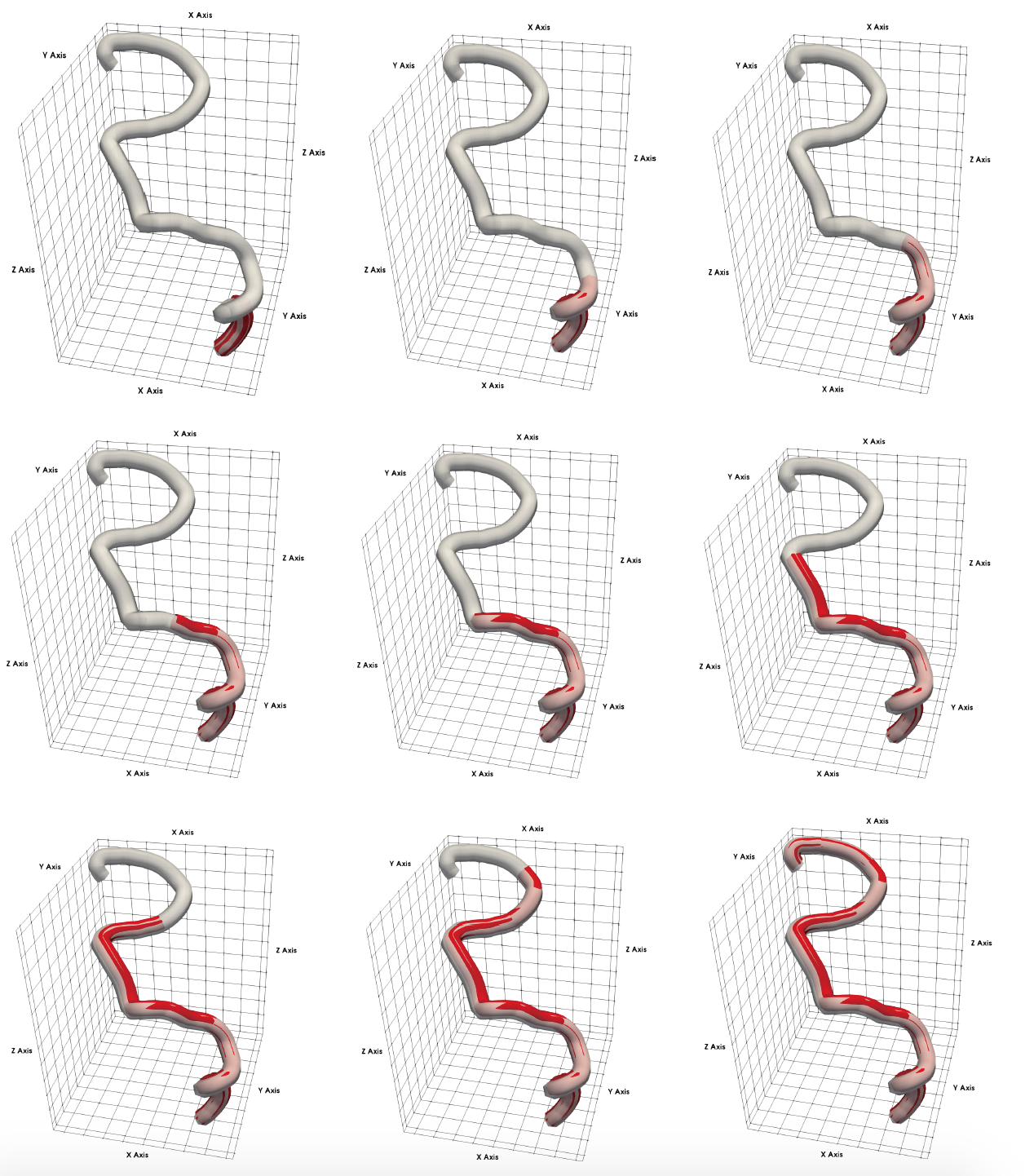}
          \caption{There are nine steps to run the trajectory controller algorithm before reaching the target path. The white path is the corridor
inside the vessel, and red line is the MRbot motion which follows the vascular trajectory closely.}
\label{nine}
\end{figure*}

\begin{figure*}[h]
\centering
\includegraphics[width = 13cm,height=14cm]{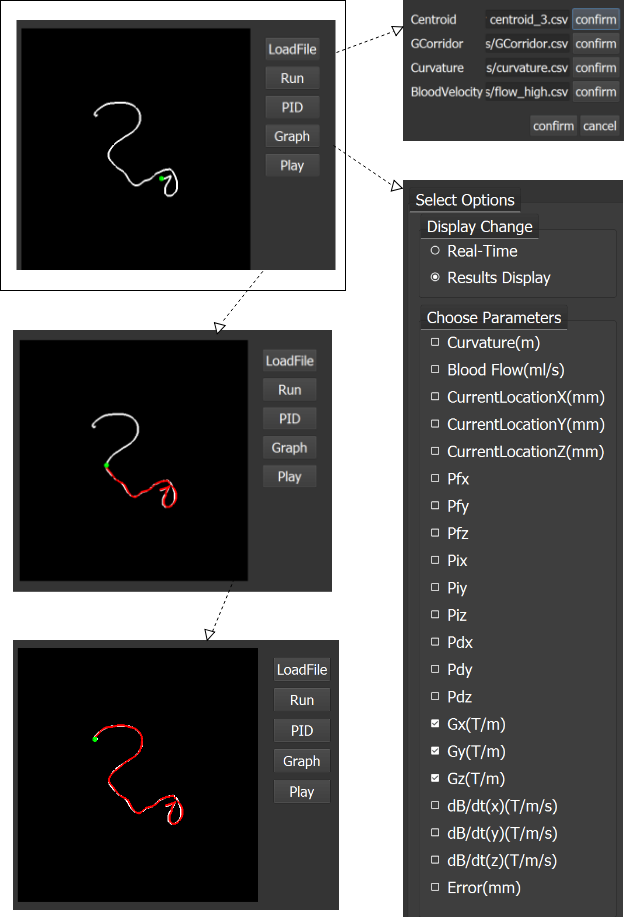}
          \caption{Structure of Qt user interface}
\label{qtt}
\end{figure*}
The built-in magnetic field strengths currently give tens of mT/m of gradients in any direction for imaging purposes. Such gradients produce a strong magnetic force in many directions. In order to correctly propel the ferromagnetic core on the pre-planned path and to provide a feel for this experiment, we measure the influence of magnetic field gradients from an MRI system on a ferromagnetic sphere. This ferromagnetic field is suitable because permendur has strong magnetization with a saturation value that is higher than other magnetic materials. In the computational core, each position has three gradients [X(i), Y(i), Z(i)] (i refers to the digitized form) on the 3D vascular path. The navigation of the ferromagnetic sphere relies mainly on two aspects: drag force D and the propulsion magnetic force F. During the experiments, robust software architecture must be present to seamlessly and precisely compute and apply the necessary gradients information. We run the experiment 1,000 times and record the results. The execute time is within 7.2ms to 8.1ms and the average running time is 7.7ms. We also calculated the spatial gradient to obtain more information. For example, in X-axis: 0.93 mT/m, 1.68 mT/m, 3.35 mT/m, and 3.038 mT/m, in Y-axis: 3.33 mT/m, 1.13 mT/m, 3.84 mT/m, and 3.53 mT/m, in Z-axis: 2.65437 mT/m, 3.7658 mT/m, 3.61 mT/m, and 2.84 mT/m. Such gradients can be used to produce a magnetic force vector in upward and downward directions. We assume blood density is constant which means that we can neither increase nor decrease the amount of blood in the vessel. The volume of blood that goes in must equal the volume that comes out.

\subsection{Time-varying Magnetic Fields}

The gradient coils are used for imaging produce time-varying magnetic fields' slew rate in dB/dt [18]. During the rise time of the magnetic field, an electric current may be induced in a conductor. The human body is a conductor which can produce peripheral nerve stimulation. This stimulation may result in a slight tingling sensation or a brief muscle twitch but is not recognized as a significant health risk [19]. In order to avoid these potential hazards, the FDA has the rule to guide gradient field switch if dB/dt is sufficient to produce propulsion in vessels. Slew rates can be calculated with [31]:

\begin{equation}
 \label{e1}
S = G_p/T_r * r
\end{equation}

where $\mathnormal{S}$ is slew rates (dB/dt), $G_p$ is the difference between two consecutive gradients, $T_r$ is rise time, and $\mathnormal{r}$ is the distance of 50 cm from the isocenter during the scan. 
dB/dt is measured in units of Tesla per meter per second (T/m/s). The slew rates and gradient strengths above are $\mathnormal{x}$-gradient, $\mathnormal{y}$-gradient, and $\mathnormal{z}$-gradient. The slew rate is limited by the FDA and is generally below 200T/m/s. Our results of slew rate are within FDA limitations. See Figure ~\ref{dC},~\ref{Da},~\ref{Dh}

\subsection{Visualization Toolkit}

Our approach is to augment an existing visualization system. The VTK is a freely available C++ class library for 3D graphics and visualization [20]-[22]. When we get each coordinate of points, we want to create a sharply focused object library that could easily embed in our sphere pathway. The sphere pathway can be built from small pieces. The key to toolkits is that pieces must be well defined, with simple interfaces. 
\subsection{Simulation Data to Image Database}

The ability to handle a large dataset is a critical requirement. From past studies, we see that ParaView is an open-source application with multi-platform visualization and scales well [23]. This tool provides a graphical user interface for the interactive exploration of large data sets. Our framework is built on top of ParaView, an analysis optimized C++ designed and modern visualization tool in post-processing, for advanced modeling and simulation codes. See Figure ~\ref {nine}.

The VTK is the foundation of the ParaView architecture which provides data representations, algorithms, and a mechanism to connect these representations and algorithms to form a working program. Red and white lines represent actual sphere maps and desired paths. The red path has 8380 experimental data points when the sphere moved along a selected path. Currently, the step size taken is 0.0001 and this small step size can be increased and determined to a value where we do not lose much accuracy. The MRbot closely follows the vascular trajectory.      
\subsubsection{Current application development platforms}
Qt is one of the most important cross-platform application frameworks that is used for the development of GUI programs. We developed web-enabled applications for deploying them across personal
computer desktops, mobile devices, and embedded operating systems
without rewriting the source code. See Figure ~\ref{qtt}. 

\section{DISCUSSION AND CONCLUDING REMARKS}
In this study, we simulated a ferromagnetic core inside a vessel by exploiting the real-time features. Our studies have shown that it is feasible to design an MRbot of this size with propulsion forces considering the characteristics of the human cardiovascular system. This tracking technique is integrated into propulsion and a PID controller successfully, which allows real-time automatic navigational control of an MRbot in the blood vessels. The core is implemented in C/C++ for streaming development, testing, and processing the output. In the work, appropriate libraries (eg. VTK) and two realistic blood flow waveforms are incorporated in this simulation.

We manipulate the MRbot with predefined paths which achieve high accuracy propulsion and transport.
However, no virtual routines that are accessible via the graphic tools of the GUI shown on the LCD and/or of a HoloLens have been done. 
Therefore, a theoretical framework to predict assisted interventions
performance is under development.


\end{document}